# INTRODUCING THE VIEWPOINT IN THE RESOURCE DESCRIPTION USING MACHINE LEARNING

Ouahiba Djama

Lire Laboratory, University of Abdelhamid
Mehri Constantine 2, Constantine, Algeria

## ABSTRACT

*Search engines allow providing the user with data and information according to their interests and specialty. Thus, it is necessary to exploit descriptions of the resources, which take into consideration viewpoints. Generally, the resource descriptions are available in RDF (e.g., DBPedia of Wikipedia content). However, these descriptions do not take into consideration viewpoints. In this paper, we propose a new approach, which allows converting a classic RDF resource description to a resource description that takes into consideration viewpoints. To detect viewpoints in the document, a machine learning technique will be exploited on an instanced ontology. This latter allows representing the viewpoint in a given domain. An experimental study shows that the conversion of the classic RDF resource description to a resource description that takes into consideration viewpoints, allows giving very relevant responses to the user's requests.*

## 1. INTRODUCTION

With the rapid increase in the amount of published information and data on the web, search engines must select the most relevant information to the user's viewpoint. Thus, the resource description should take into consideration the different viewpoints of users. However, the existing descriptions on the web do not take into consideration the notion of the viewpoint. For that, we aim to convert the existing descriptions to descriptions that take into consideration the different viewpoints of users instead of building new descriptions.

Generally, on the web, the resource descriptions are available in RDF as DBPedia of Wikipedia content, etc. Djama [1] has proposed an RDF based framework (VP-RDF) to introduce the viewpoint in the description of resources.

Currently, no tool allows converting the existing RDF documents to VP-RDF documents. The conversion of the RDF document to VP-RDF document, allows introducing the viewpoint in the existing resource description.

For that, in this paper, we aim to propose an approach that allows converting RDF document to VP-RDF document.

The RDF document belongs to a given domain. Thus, we will apply the machine learning technique on the instantiated multi-viewpoints ontology [2] of this domain to detect the relations between concepts / instances and viewpoints.

In the following section, we present a state of art. Then, in Section 3, we explain the proposed approach. After that, in Section 4 we apply the proposed approach to some use cases. Section 5 provides some results of the research. Finally, we present some areas for future work.

## 2. STATE OF THE ART

In this section, we present the definition of the viewpoint, some related work on the viewpoint and the machine learning with the ontology, VP-RDF language and the instantiated multi-viewpoints ontology.

### 2.1. Viewpoint Definitions

The authors have proposed different definitions of the notion of the viewpoint. In some works, the viewpoint corresponds to the perception of an object according to the observer's position [3]. For example, according to the observer's positions about the symbol '9', there are two viewpoints: number 6 and number 9 [2].

In some other work, (e.g., [1], [2], [4], [5] and [6]), the viewpoint is defined as a partial definition of an object basing on only some set of properties of this object. For example, in [2], the object apartment is defined by the size properties as area, room number, height, etc. and the finance properties as rent, price, etc. Therefore, we can give two different descriptions of the same apartment:

1) Viewpoint 1: large apartment, according to viewpoint 'size'.
2) Viewpoint 2: expensive apartment, according to viewpoint 'finance'.

Remark: Someone describes an apartment as cheap apartment and another one describes it as expensive apartment. This case is related to the fuzzy notion and not the viewpoint.

For example, the user's request aim to find all the existing properties of an apartment that describe its size [1]. With the exploitation of the classic descriptions, the search engine gives all the properties of this apartment [1] because it cannot detect that such properties are linked to the viewpoint size. This latter is because; the existing resource descriptions do not show the relations between the properties and the viewpoint [1].

The resource description that takes into consideration the viewpoint, allows linking each resource (property or entity) to a viewpoint [1].

### 2.2. Related Work

Several work are interested in the notion of the viewpoint. The authors in [5], [7], [8], [9], [10] and [11] have integrated the viewpoint in the development of the ontology. This ontology called 'multi-viewpoints ontology'.

Djezzar and Boufaida [12] have proposed an approach of the classification of an individual in the multi-viewpoints ontology. Djakhdjakha et al. [13] are interested in the alignment of the multi-viewpoints ontologies. Djama and Boufaida [2] have developed an approach that allows instancing the multi-viewpoints ontology.

These works are interested in the treatment of the ontology with viewpoints and not the treatment of the resources and documents with viewpoints.

Djama and Boufaida have also proposed an approach, in [14] and [15], which allows using multi-viewpoints ontology to annotate resources. This annotation allows describing the resource elements using the ontology elements. Then, the obtained annotation can be represented in RDF. Djama [1] has proposed the VP-RDF as an extension of the RDF to introduce the viewpoint in the description of resources.

Therefore, the works [1], [2], [14] and [15] allows constructing an RDF document from informal document as text document, XML document, etc. However, we aim to introduce the viewpoint in the existing RDF document. We will not construct an RDF document, but we will convert an existing RDF document to VP-RDF document.

Martin *et al.* [16] have exploited the viewpoints for reasoning on the classical ontology using case-based approach. In this work, the authors have not integrated the viewpoint in the ontology. Thus, the different viewpoints do not belong to a given domain. However, in our approach, the different viewpoints should appear in a given domain.

Gorshkov *et al.* [17] have exploited a multi-viewpoints ontology as a decision-making support. In this work, the authors have exploited the viewpoints that are represented in the ontology to make a decision. However, in our work, we aim to exploit the viewpoints that are represented in the instantiated multi-viewpoints ontology to recognize the set of the viewpoints in a given domain. These viewpoints will be introduced in the existing RDF documents. Thus, the resource description will be enriched.

Trichet *et al.* [18] have introduced the viewpoints in the semantic annotation of images. The authors have developed a platform that allows a user to use a set of ontologies to create a semantic annotation according to his/her viewpoint. The semantic annotation will be represented in RDF. However, the notion of the viewpoint cannot appear clearly in the RDF representation [1]. In the VP-RDF [1], the viewpoint appears clearly. Therefore, in our work, we aim to propose an approach that allows converting RDF document to VP-RDF document.

Several works allow introducing the context in RDF, as in [19], [20], [21] and [23]. These works allow representing that an assertion is true under a given context. For example [1], parallel lines can intersect in the context of solid geometry (3D geometry). However, in our work, we aim to represent the relation between a description of a resource and a viewpoint. For example, this is a large apartment; according to viewpoint 'size' and it is an expensive apartment, according to viewpoint 'finance'.

According to Djama [1], the context and the viewpoint are two different notions. The context is a judgment based on rational arguments that represent a set of conditions [1] (Euclidean geometry or solid geometry). However, the viewpoint is a partial definition of an object [1].

Doan *et al.* [24] have developed a machine learning approach to establish semantic mappings among multiple ontologies. This approach is based on well-founded notions of semantic similarity, expressed in terms of the joint probability distribution of the concepts involved. The authors described the use of multi-strategy learning for computing concept similarities.

Kulmanov *et al.* [25] provided an overview over the methods that use ontologies to compute similarity and incorporate them in machine learning methods. The authors outline how semantic similarity measures and ontology embeddings can exploit the background knowledge in ontologies and how ontologies can provide constraints that improve machine-learning models.

The works [24] and [25] allow using machine-learning techniques to compute similarity between ontology concepts. However, we aim to make predictions of the relations between concepts/instances with the viewpoints.

## 2.3. VP-RDF

According to Djama [1], VP-RDF is an extension of the RDF by adding a new type of statement. The latter is composed of (Subject, Predicate_with_Viewpoint, Viewpoint) [1]. This statement allows linking a resource (Subject) to a viewpoint via the predicate Predicate_with_Viewpoint. To create this statement, Djama [1] proposed new elements that are shown in the table 1 and the table 2.

Table 1. VP-RDF Classes

| Class | Definition by the RDF vocabulary |
|---|---|
| VPrdf:Viewpoint | Subclass of "rdfs:Resource" |
| VPrdf:Predicate_with_Viewpoint | Subclass of "rdf:Property" |
| VPrdf:Statement | Subclass of "rdf:Statement" |

Table 2. VP-RDF Properties

| Property | Domain | Range |
|---|---|---|
| VPrdf:Subject_Statement | VPrdf:Statement | rdfs:Resources |
| VPrdf:Predicate_Statement | VPrdf:Statement | VPrdf:Predicate_with_Viewpoint |
| VPrdf:Object_Statement | VPrdf:Statement | VPrdf:Viewpoint |

## 2.4. Instantiated Multi-Viewpoints Ontology

The multi-viewpoints ontology [8] allows representing domain knowledge in two levels: Consensual and Heterogeneous level.

The consensual level allows representing the consensual concepts that are defined in all the viewpoints in the domain [2]. So, the consensual concepts can be considered not linked to particular viewpoints. These concepts are called global concept. Each of them is defined by a set of global properties (attributes) and a set of local properties [2]. A global property is defined in all the viewpoints [2]. So, a global property is consensual and can be considered not linked to particular viewpoints. Each local property is defined according to some viewpoints [2]. So, this property is linked to these viewpoints. The global concepts are organized in a global hierarchy [2].

For example [2], in the real estate domain, the global concept *Apartment* is defined by the properties: *surface, number of rooms, height*, etc. according to the viewpoint *Size*. The global concept *Apartment* is defined also by the properties: *price, taxes, rent price*, etc. according to the viewpoint *Finance*.

The value of a local property of a global concept allows generating a local concept according to the same viewpoint where this property is defined [2]. So, this concept is linked to this viewpoint. On the other hand, the global concept subsumes the local concept [2]. From this local concept, a hierarchy of local concepts will be built in the same viewpoint where this local concept is defined [2]. This local hierarchy of local concepts represents a local representation according to this viewpoint. The same global concept can be defined also by another local property that is defined in another viewpoint. In the same way, another local representation according of another viewpoint will be generated. And so on, the set of local representations represents the heterogeneous level.

For example [2], according to the property *surface*, we can generate the local concept *Large apartment*. This latter is defined according to the viewpoint *Size*. According to the property *price*, we can generate the local concept *Expensive apartment*. This latter is defined according to the viewpoint *Finance*. The global concept *Apartment* subsumes the local concept *Large apartment* and the local concept *Expensive apartment.*

The relationships between local concepts of the same viewpoint are called local roles [2]. Each relation is linked to the viewpoint where the local concepts are linked to. The relationships between local concepts of two different viewpoints are called global roles [2]. The global roles are not linked to particular viewpoints.

For example [2], the local role *rent* allows linking the local concepts *Rich tenant* and *expensive apartment* that are defined according to the viewpoint *Finance*. The local role *lives* allows linking the local concepts *Rich tenant* and *Large apartment. Rich tenant* is defined according to the viewpoint *Finance*. However, *Large apartment* is defined according to the viewpoint *size*.

The figure 1 shows the multi-viewpoints ontology structure.

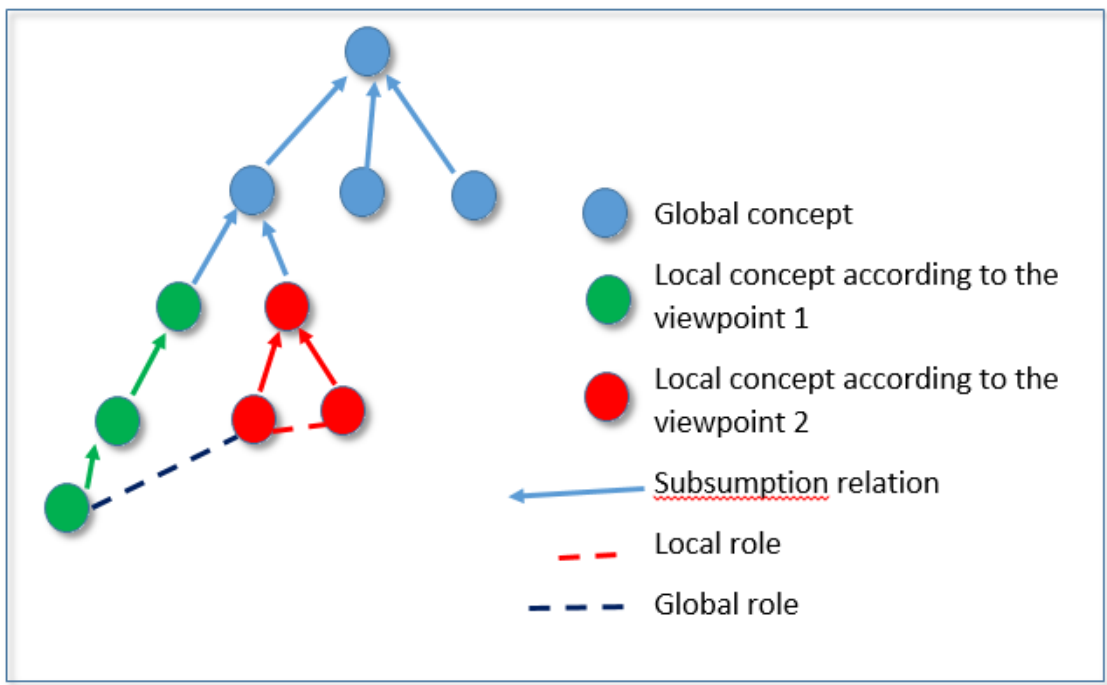

Figure 1.  The multi-viewpoints ontology structure

The instantiated multi-viewpoints ontology is a multi-viewpoints ontology that contains instances. An individual can be an instance of only one local concept under a particular viewpoint [2]. This individual can be an instance of another local concept in another viewpoint [2]. This individual is an instance of the global concept that subsumes all these local concepts.

The individual *apartment N°1* is an instance of the local concept *Large apartment* under the viewpoint *Size* and the local concept *Expensive apartment* under the viewpoint *Finance*. The individual *apartment N°1* is an instance of the global concept *Apartment.*

An individual, that is an instance of a local concept under a particular viewpoint, is linked to this viewpoint.

An individual, that is an instance of only a global concept, is not linked to particular viewpoints. Therefore, the instantiated multi-viewpoints ontology allows representing knowledge in two levels. The consensual knowledge, which are not linked to viewpoints in the domain, are represented in the consensual level. The knowledge, which are linked to viewpoints, are represented in the heterogeneous level. We resume the descriptions of the components of the instantiated multi-viewpoints ontology in the table 3.

Table 3. Components of the instantiated multi-viewpoints ontology

| Element | Description |
|---|---|
| Global concept | Not linked to viewpoints |
| Global attribute (data type property) | Not linked to viewpoints |
| Global role (object property) | Not linked to viewpoints |
| Global instance | Not linked to viewpoints |
| Local concept | Linked to one or several viewpoints |
| Local attribute (data type property) | Linked to one or several viewpoints |
| Local role (object property) | Linked to one or several viewpoints |
| Local instance | Linked to one or several viewpoints |

## 3. PROPOSED APPROACH

The proposed approach aims to convert an RDF document to VP-RDF document. For that, it is composed of two main steps:

### 3.1. Detection of Viewpoints

An RDF document is composed of a set of RDF triplets (statements). Each statement is composed of subject, predicate and object. This step aims to detect for each statement and for each its component the viewpoint that is linked to. There are four main cases:

1) Subject of the statement is linked to one or several viewpoints.
2) Object of the statement is linked to one or several viewpoints.
3) Both the subject and the object are linked to one or several viewpoints.
4) Predicate of the statement is linked to one or several viewpoints.

The other subcases are the combination between these cases.

In this step, we aim to make predictions:

a) which viewpoint that the RDF subject is linked to,
b) which viewpoint that the RDF object is linked to,
c) which viewpoint that the RDF predicate is linked to.

To make predictions, it is necessary to exploit a machine learning techniques. The RDF document belongs to a given domain. Thus, we will exploit the instantiated multi-viewpoints ontologies [2] of this domain to extract viewpoints in the domain. Therefore, we create a model that will learn, from various instantiated multi-viewpoints ontologies of the same domains, to link a term to a viewpoint. A term can be a concept or an instance of a concept. This model will learn also to link a relation between terms to a viewpoint. Therefore, we can exploit a machine learning technique that is based on the computing of the frequencies of relations between terms and viewpoints.

After the machine-learning step, the created model is able to predict (detect) the viewpoint that the RDF subject / the RDF object / the RDF predicate is linked to. We choose to use a machine learning technique to detect a viewpoint in order to avoid reasoning on the instantiated multi-viewpoints ontologies, each time we want to convert an RDF document to a VP-RDF document. The reasoning on the ontologies takes a considerable amount of time.

Now, the created model will be exploited in the first step in our approach to detect:

1) The viewpoint/ the set of viewpoints where the subject of the statement is linked to.
2) The viewpoint/ the set of viewpoints where the object of the statement is linked to.
3) The viewpoint/ the set of viewpoints where the predicate of the statement is linked to.

For example, in the real estate domain, *Rich_Tenant* is always linked to the viewpoint *Finance*. *Large_Apartment* is linked to the viewpoint *size*. The relation *rent* is linked to the viewpoint *Finance*.

For example, in the education domain, *Professor* is linked to the viewpoint *University_Education*.

## 3.2. Creation of VP-RDF Statements

This step aims to exploit the VP-RDF vocabulary and syntax [1] (see subsection 2.3.) to convert the RDF statement to VP-RDF statement. There are four cases:

1. Subject of the statement is linked to one or several viewpoints: in this case, the RDF statement will be kept and from the subject, one or several VP-RDF statements will be created. For example, the RDF statement (<Rich_Tenant>,<lives_in>,<Constantine>) will be converted in VP-RDF in two statements:
   (<Rich_Tenant>,<lives_in>,<Constantine>) and
   (<Rich_Tenant>,<belong_to>,<Finance>)).
   From the subject 'Rich_Tenant', one VP-RDF statement will be created; because 'Rich_Tenant' is linked to only one viewpoint.
2. Object of the statement is linked to one or several viewpoints: in this case, the RDF statement will be kept and from the object, one or several VP-RDF statements will be created.
3. Both the subject and the object are linked to one or several viewpoints: in this case, the RDF statement will be kept. From the subject, one or several VP-RDF statements will be created. From the object, one or several VP-RDF statements will be created.

4.   Predicate of the statement is linked to one or several viewpoints: in this case, first, it is necessary to create a class to represent the predicate. Then, this class will be linked to a viewpoint via a VP-RDF statement. It will be linked also to the object of the RDF statement via a new RDF statement, where this class becomes a subject of the new RDF statement.

The figure 2 shows the creation of the machine-learning model. The figure 3 shows the different steps of the proposed approach with the inputs and outputs.

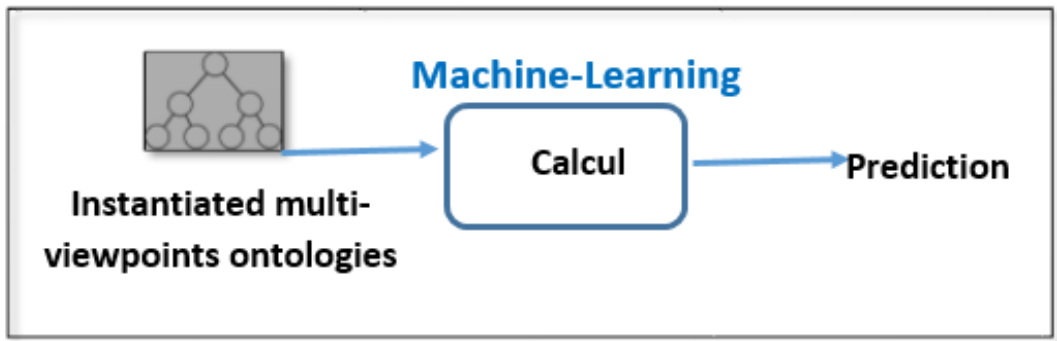

Figure 2.  The creation of the machine-learning model

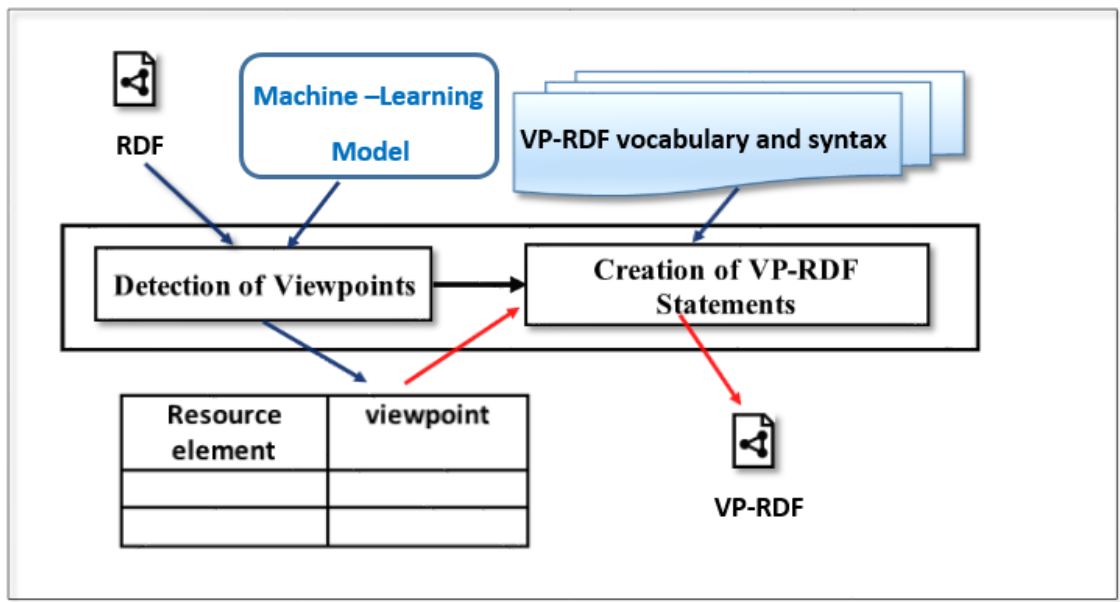

Figure 3.  Schema of the proposed approach

## 4.  USE CASES

In this section, we present an example for each case seen in the previous section. These examples are regrouped in the following table:

Table 4.  Use Cases

| Statement in RDF Document | Conversion to VP-RDF |
|---|---|
| (<Rich_Tenant>,<lives_in>,<Constantine>) | (<Rich_Tenant>, <lives_in>, <Constantine>) <br><br> (<Rich_Tenant >, < belong_to>, <Finance>) |
| (<John>,<is>, <Professor >) | (<John>, <is>, <Professor >) <br><br> (<Professor > <defined_according _to>, <University_Education >) |
| (<Rich_Tenant>,<lives_in>,<Large_Apart ment>) | (<Rich_Tenant>, <lives_in>, <Large_Apartment>) <br><br> (<Rich_Tenant >, < belong_to>, <Finance>) <br><br> (<Large_Apartment> <according_to> <size>) |
| (<John>,<rent>, <apartment3>) | (<John>,<rent>, <apartment3>) <br><br> (<Class_rent >, < rent _value>, < apartment3>) <br><br> (<Class_rent >, < belong_to>, <Finance>) |

In this table, the statements that are written with red colour represent VP-RDF statements in the VP-RDF language. The black ones represent RDF statements in the VP-RDF language.

## 5. RESULTS AND DISCUSSION

We have realized an experimental study to evaluate the effect of the exploitation of the VP-RDF documents in the search of responses to the user's requests. We have converted some pre-existed RDF documents to VP-RDF documents. The RDF documents belong to different domains: 150 RDF documents of the real estate domain, 125 RDF documents of the library domain, 175 RDF documents of the education domain and 1200 RDF documents of the medical domain.

We asked 200 users to make a request for each these domains according to their interests (viewpoints). We exploited the VP-RDF documents to give responses to the user's requests. We asked the same users to evaluate these responses if are relevant to their interests. Finally, we calculated the percentage of the number of responses that are relevant to the user's interest. The results are regrouped into the table 5:

Table 5.  Results of the study

| Domain | Percentage of the Relevant Responses |
|---|---|
| Real estate | 98.1% |
| Library | 98.3% |
| Education | 98% |
| Medical | 94.4% |

We find that all the percentages are high in all the domains. This is because the relationships between viewpoints and resources are explicitly represented in the VP-RDF documents.

For example, in the real estate domain, a user wants to know all the existing properties of an apartment that describe its size [1].

In VP-RDF, the properties (e.g., height, *surface*, the number of rooms, etc.) are linked directly (explicitly) to the viewpoint size [1]. The answer shows the values of height, *surface*, the number of rooms, etc. So, the answer will be found directly without reasoning. However, in the RDF, the properties (e.g., height, *surface*, the number of rooms, etc.) are not linked to the viewpoint size. The answer shows the values of the all properties of the apartment. In order to improve the answer, it is necessary to exploit a reasoning tool to select only the values of the properties: height, *surface*, the number of rooms, etc. However, the reasoning can give the results that are not accurate. For example, the answer can be the value of the properties: *height* and *surface*. However, *number of rooms* will not be mentioned. Therefore, with classic RDF, the answer can be incomplete or more than the user's needs. In the last case, the user should select the relevant answer manually.

With VP-RDF, the answer will be completely relevant exactly to the user's needs. Therefore, the user does not need to select the relevant answer manually.

## 6. CONCLUSIONS

We have proposed an approach that allows converting RDF document to VP-RDF document. This latter allows introducing explicitly the viewpoint in the description of the resources and their relationships [1]. This description helps search engines to give responses that are relevant to the user's interest (viewpoint) with high rate and the user does not need to select the relevant answer manually. The machine learning mechanism is exploited to detect viewpoints basing on the instantiated multi-viewpoints ontology.

The accuracy of the proposed method depends on the accuracy of the detection of the viewpoints in a given domain. It depends also on the accuracy of the predictions about the relation between resources and viewpoints. Therefore, the amount and the quality of the data exploited in the machine learning affect the accuracy of the proposed approach.

The imprecision and the uncertainty can be coupled with the viewpoints in the resource description. For example, in the finance viewpoint, someone describes an apartment as cheap apartment and another one describes it as expensive apartment. This case, the two persons share the same viewpoint (Finance). In addition to this, the descriptions are related to the fuzzy notion too. As future work, we plan to couple the viewpoint with the fuzzy in the description of resources.

## AUTHOR

**Ouahiba Djama** was born in Constantine, Algeria. She received her Engineer degree in Computer Science
in 2005 and M. sc. in 2010, both from University of Mentouri, Constantine, Algeria. She obtained a
Doctoral degree in Computer Science in 2020 from the University of Abdelhamid Mehri Constantine 2, Ali
Mendjeli, Constantine, Algeria. She is currently an Assistant Professor at the University of Mentouri
Brothers-Constantine 1, Constantine, Algeria. She is also an Attached Member of the SI&BC research
group at Lire Laboratory of the University of Abdelhamid Mehri-Constantine 2, Constantine, Algeria. Her
research interests include knowledge representation and reasoning, formal knowledge representation for
semantic web, ontology development, web technologies, big data, bioinformatics, artificial intelligence and
computer applications.
djama_ouahiba@umc.edu.dz